\documentclass[11pt,a4paper]{article}
\usepackage[hyperref]{acl2021}
\usepackage{times}
\usepackage{latexsym}
\usepackage{paralist}
\usepackage{longtable}

\newcommand{\Sref}[1]{\S\ref{#1}}
\newcommand{\Fref}[1]{Figure~\ref{#1}}
\newcommand{\Tref}[1]{Table~\ref{#1}}
\newcommand{\Aref}[1]{Appendix~\ref{#1}}

\usepackage{microtype}
\usepackage{graphicx}
\usepackage{csquotes}

\urldef{\censusurl}\url{https://www.census.gov/mso/www/training/pdf/race-ethnicity-onepager.pdf/}
\urldef{\censusurlv}\url{https://www.census.gov/topics/population/race/about.html}
\urldef{\pewurl}\url{https://www.pewresearch.org/fact-tank/2015/06/15/is-being-hispanic-a-matter-of-race-ethnicity-or-both/}

\aclfinalcopy 
\def\aclpaperid{2863} 

\newcommand\AAE{{AAE }}

\title{A Survey of Race, Racism, and Anti-Racism in NLP}

\author{Anjalie Field \\
   Carnegie Mellon University \\
  \texttt{anjalief@cs.cmu.edu} \\\And
  Su Lin Blodgett \\
  Microsoft Research  \\
  \texttt{sulin.blodgett@microsoft.com} \\\AND
  Zeerak Waseem \\
  University of Sheffield \\
  \texttt{z.w.butt@sheffield.ac.uk} \\\And
  Yulia Tsvetkov\\
  University of Washington \\
  \texttt{yuliats@cs.washington.edu}
  }

\date{}

\begin{document}
\maketitle
\begin{abstract}
Despite inextricable ties between race and language, little work has considered race in NLP research and development. In this work, we survey $79$ papers from the ACL anthology that mention race. These papers reveal various types of race-related bias in all stages of NLP model development, highlighting the need for proactive consideration of how NLP systems can uphold racial hierarchies. However, persistent gaps in research on race and NLP remain: race has been siloed as a niche topic and remains ignored in many NLP tasks; most work operationalizes race as a fixed single-dimensional variable with a ground-truth label, which risks reinforcing differences produced by historical racism; and the voices of historically marginalized people are nearly absent in NLP literature. By identifying where and how NLP literature has and has not considered race, especially in comparison to related fields, our work calls for inclusion and racial justice in NLP research practices.
\end{abstract}

\section{Introduction}
\label{sec:intro}

Race and language are tied in complicated ways. Raciolinguistics scholars have studied how they are mutually constructed: historically, colonial powers construct linguistic and racial hierarchies to justify violence, and currently, beliefs about the inferiority of racialized people's language practices continue to justify social and economic exclusion \cite{rosa_flores_2017}.\footnote{We use \textit{racialization} to refer the process of ``ascribing and prescribing a racial category or classification to an individual or group of people \ldots based on racial attributes including but not limited to cultural and social history, physical features, and skin color'' \cite{hudley2017language}.}
Furthermore, language is the primary means through which stereotypes and prejudices are communicated and perpetuated \cite{hamilton1986stereotypes,bar2013stereotyping}.

However, questions of race and racial bias have been minimally explored in NLP literature. While researchers and activists have increasingly drawn attention to racism in computer science and academia, frequently-cited examples of racial bias in AI are often drawn from disciplines other than NLP, such as computer vision (facial recognition) \citep{buolamwini2018gender} or machine learning (recidivism risk prediction) \cite{angwin2019machine}. Even the presence of racial biases in search engines like Google \cite{sweeney2013discrimination,noble2018algorithms} has prompted little investigation in the ACL community. Work on NLP and race remains sparse, particularly in contrast to concerns about gender bias, which have led to surveys, workshops, and shared tasks \citep{sun-etal-2019-mitigating,webster-etal-2019-gendered}.

In this work, we conduct a comprehensive survey of how NLP literature and research practices engage with race. We first examine $79$ papers from the ACL Anthology that mention the words `race', `racial', or `racism' and highlight examples of how racial biases manifest at all stages of NLP model pipelines (\Sref{sec:survey}). We then describe some of the limitations of current work (\Sref{sec:limitations}), specifically showing that NLP research has only examined race in a narrow range of tasks with limited or no social context. Finally, in \Sref{sec:people}, we revisit the NLP pipeline with a focus on how \textit{people} generate data, build models, and are affected by deployed systems, and we highlight current failures to engage with people traditionally underrepresented in STEM and academia. 

While little work has examined the role of race in NLP specifically, prior work has discussed race in related fields, including human-computer interaction (HCI) \citep{Ogbonnaya-Ogburu2020,Rankin2019,Schlesinger2017}, fairness in machine learning \cite{hanna2020towards}, and linguistics \cite{hudley2020toward,motha_2020}. We draw comparisons and guidance from this work and show its relevance to NLP research. Our work differs from NLP-focused related work on gender bias \cite{sun-etal-2019-mitigating}, `bias' generally \cite{blodgett-etal-2020-language}, and the adverse impacts of language models \cite{bender-2021} in its explicit focus on race and racism.

In surveying research in NLP and related fields, we ultimately find that \textit{NLP systems and research practices produce differences along racialized lines}. Our work calls for NLP researchers to consider the social hierarchies upheld and exacerbated by NLP research and to shift the field toward ``greater inclusion and racial justice'' \cite{hudley2020toward}.

\section{What is race?}
\label{sec:race_definition}
It has been widely accepted by social scientists that race is a social construct, meaning it ``was brought
into existence or shaped by historical events, social forces, political power, and/or colonial conquest'' rather than reflecting biological or `natural' differences \cite{hanna2020towards}. More recent work has criticized the ``social construction'' theory as circular and rooted in academic discourse, and instead referred to race as ``colonial constituted practices'', including ``an inherited western, modern-colonial practice of violence, assemblage, superordination, exploitation and segregation'' \citep{saucier2016conceptual}.

The term race is also multi-dimensional and can refer to a variety of different perspectives, including \emph{racial identity} (how you self-identify), \emph{observed race} (the race others perceive you to be), and \emph{reflected race} (the race you believe others perceive you to be)  \cite{roth2016multiple, hanna2020towards, Ogbonnaya-Ogburu2020}. Racial categorizations often differ across dimensions and depend on the defined categorization schema. For example, the United States census considers Hispanic an ethnicity, not a race, but surveys suggest that 2/3 of people who identify as Hispanic consider it a part of their racial background.\footnote{\censusurl, \censusurlv, \pewurl} Similarly, the census does not consider `Jewish' a race, but some NLP work considers anti-Semitism a form of racism \cite{hasanuzzaman-etal-2017-demographic}. Race depends on historical and social context---there are no `ground truth' labels or categories \cite{roth2016multiple}.

As the work we survey primarily focuses on the United States, our analysis similarly focuses on the U.S.
However, as race and racism are global constructs, some aspects of our analysis are applicable to other contexts. We suggest that future studies on racialization in NLP ground their analysis in the appropriate geo-cultural context, which may result in findings or analyses that differ from our work.

\begin{table*}[ht]
    \centering
    \begin{tabular}{lcccccc|c}
    & \rotatebox[origin=l]{90}{\small{Collect Corpus}} & \rotatebox[origin=l]{90}{\small{Analyze Corpus}} & \rotatebox[origin=l]{90}{\small{Develop Model}} & \rotatebox[origin=l]{90}{\small{Detect Bias}} & \rotatebox[origin=l]{90}{\small{Debias}} & \rotatebox[origin=l]{90}{\small{Survey/Position}} & \rotatebox[origin=l]{90}{\small{Total}} \\
 \hline
 \hline

Abusive Language & 6 & 4 & 2 & 5 & 2 & 2 & 21 \\
Social Science/Social Media & 2 & 10 & 6 & 1 & - & 1 & 20 \\
Text Representations \small{(LMs, embeddings)} & - & 2 & - & 9 & 2 & - & 13 \\
Text Generation \small{(dialogue, image captions, story gen.
)}  & - & - & 1 & 5 & 1 & 1 & 8 \\
Sector-specific NLP applications \small{(edu., law, health)} & 1 & 2 & - & - & 1 & 3 & 7 \\
Ethics/Task-independent Bias & 1 & - & 1 & 1 & 1 & 2 & 6 \\
Core NLP Applications \small{(parsing, NLI, IE)} & 1 & - & 1 & 1 & 1 & - & 4 \\
\hline
Total & 11 & 18 & 11 & 22 & 8 & 9 & 79 \\
    \end{tabular}
    \caption{$79$ papers on race or racism from the ACL anthology, categorized by NLP application and focal task.}
    \label{tab:task_overview}
\end{table*}

\section{Survey of NLP literature on race}
\label{sec:survey}

\subsection{ACL Anthology papers about race}
In this section, we introduce our primary survey data---papers from the \href{https://www.aclweb.org/anthology/}{ACL Anthology}\footnote{The ACL Anthology includes papers from all official ACL venues and some non-ACL events listed in \Aref{sec:appendix_venues}, as of December $2020$ it included $6,200$ papers}---and we describe some of their major findings to emphasize that \textit{NLP systems encode racial biases}. We searched the anthology for papers containing the terms `racial', `racism', or `race', discarding ones that only mentioned race in the references section or in data examples and adding related papers cited by the initial set if they were also in the ACL Anthology. In using keyword searches, we focus on papers that explicitly mention race and consider papers that use euphemistic terms to not have substantial engagement on this topic. As our focus is on NLP and the ACL community, we do not include NLP-related papers published in other venues in the reported metrics (e.g.~\Tref{tab:task_overview}), but we do draw from them throughout our analysis.

Our initial search identified $165$ papers.
However, reviewing all of them revealed that many do not deeply engage on the topic. For example, $37$ papers mention `racism' as a form of abusive language or use `racist' as an offensive/hate speech label without further engagement. $30$ papers only mention race as future work, related work, or motivation, e.g.~in a survey about gender bias, ``Non-binary genders as well as racial biases have largely been ignored in NLP'' \citep{sun-etal-2019-mitigating}. After discarding these types of papers, our final analysis set consists of $79$ papers.\footnote{We do not discard all papers about abusive language, only ones that exclusively use racism/racist as a classification label. We retain papers with further engagement, e.g. discussions of how to define racism or identification of racial bias in hate speech classifiers.}

\Tref{tab:task_overview} provides an overview of the $79$ papers, manually coded for each paper's primary NLP task and its focal goal or contribution. We determined task/application labels through an iterative process: listing the main focus of each paper and then collapsing similar categories. In cases where papers could rightfully be included in multiple categories, we assign them to the best-matching one based on stated contributions and the percentage of the paper devoted to each possible category. In the Appendix we provide additional categorizations of the papers according to publication year, venue, and racial categories used, as well as the full list of $79$ papers.

\subsection{NLP systems encode racial bias}
\label{sec:nlp_pipeline}

Next, we present examples that identify racial bias in NLP models, focusing on $5$ parts of a standard NLP pipeline: data, data labels, models, model outputs, and social analyses of outputs. 
We include papers described in \Tref{tab:task_overview} and also relevant literature beyond the ACL Anthology (e.g. NeurIPS, PNAS, Science).
These examples are not intended to be exhaustive, and in \Sref{sec:limitations} we describe some of the ways that NLP literature has failed to engage with race, but nevertheless, we present them as evidence that \textit{NLP systems perpetuate harmful biases along racialized lines}.

\paragraph{Data}
A substantial amount of prior work has already shown how NLP systems, especially word embeddings and language models, can absorb and amplify social biases in data sets \cite{Bolukbasi2016ManIT,zhao-etal-2017-men}.
While most work focuses on gender bias, some work has made similar observations about racial bias \cite{rudinger-etal-2017-social,garg2018word,kurita-etal-2019-measuring}. These studies focus on \textit{how} training data might describe racial minorities in biased ways, for example, by examining words associated with terms like `black' or traditionally European/African American names \cite{Caliskan183,manzini-etal-2019-black}. Some studies additionally capture \textit{who} is described, revealing under-representation in training data, sometimes tangentially to primary research questions: \citet{rudinger-etal-2017-social} suggest that gender bias may be easier to identify than racial or ethnic bias in Natural Language Inference data sets because of data sparsity, and \citet{Caliskan183} alter the Implicit Association Test stimuli that they use to measure biases in word embeddings because some African American names were not frequent enough in their corpora.

An equally important consideration, in addition to whom the data describes is \textit{who authored the data}. For example, \citet{blodgett-etal-2018-twitter} show that parsing systems trained on White Mainstream American English perform poorly on African American English (AAE).\footnote{We note that conceptualizations of AAE and the accompanying terminology for the variety have shifted considerably in the last half century; see \citet{king2020from} for an overview.} In a more general example, Wikipedia has become a popular data source for many NLP tasks. However, surveys suggest that Wikipedia editors are primarily from white-majority countries,\footnote{\url{https://meta.wikimedia.org/wiki/Research:Wikipedia_Editors_Survey_2011_April}} and several initiatives have pointed out systemic racial biases in Wikipedia coverage \citep{Adams2019,Field2021Wiki}.\footnote{\url{https://en.wikipedia.org/wiki/Racial_bias_on_Wikipedia}} Models trained on these data only learn to process the type of text generated by these users, and further, only learn information about the topics these users are interested in. The \textit{representativeness} of data sets is a well-discussed issue in social-oriented tasks, like inferring public opinion \citep{olteanu2019social}, but this issue is also an important consideration in `neutral' tasks like parsing \cite{waseem2021disembodied}. The type of data that researchers choose to train their models on does not just affect what \textit{data} the models perform well for, it affects what \textit{people} the models work for. NLP researchers cannot assume models will be useful or function for marginalized people unless they are trained on data generated by them.

\paragraph{Data Labels}
Although model biases are often blamed on raw data, several of the papers we survey identify biases in the way researchers categorize or obtain data annotations. For example:

\begin{compactitem}
    \item \textbf{Annotation schema} Returning to \citet{blodgett-etal-2018-twitter}, this work defines new parsing standards for formalisms common in AAE, demonstrating how parsing labels themselves were not designed for racialized language varieties.
    \item \textbf{Annotation instructions} \citet{sap-etal-2019-risk} show that annotators are less likely to label tweets using \AAE as offensive if they are told the likely language varieties of the tweets. Thus, how annotation schemes are designed (e.g.~what contextual information is provided) can impact annotators' decisions, and failing to provide sufficient context can result in racial biases.
    \item \textbf{Annotator selection} \citet{waseem-2016-racist} show that feminist/anti-racist activists assign different offensive language labels to tweets than figure-eight workers, demonstrating that annotators' lived experiences affect data annotations.
\end{compactitem}

\paragraph{Models}
Some papers have found evidence that model instances or architectures can change the racial biases of outputs produced by the model. \citet{sommerauer-fokkens-2019-conceptual}  find that the word embedding associations around words like `race' and `racial' change not only depending on the model architecture used to train embeddings, but also on the specific model \textit{instance} used to extract them, perhaps because of differing random seeds. \citet{kiritchenko-mohammad-2018-examining} examine gender and race biases in $200$ sentiment analysis systems submitted to a shared task and find different levels of bias in different systems. As the training data for the shared task was standardized, all models were trained on the same data. However, participants could have used external training data or pre-trained embeddings, so a more detailed investigation of results is needed to ascertain which factors most contribute to disparate performance.

\paragraph{Model Outputs}

Several papers focus on model outcomes, and how NLP systems could perpetuate and amplify bias if they are deployed:

\begin{compactitem}
    \item Classifiers trained on common abusive language data sets are more likely to label tweets containing characteristics of \AAE as offensive \cite{davidson-etal-2019-racial,sap-etal-2019-risk}.
    \item Classifiers for abusive language are more likely to label text containing identity terms like `black' as offensive \cite{dixon2018measuring}.
    \item GPT outputs text with more negative sentiment when prompted with \AAE-like inputs  \cite{groenwold-etal-2020-investigating}.
\end{compactitem}

\paragraph{Social Analyses of Outputs}
While the examples in this section primarily focus on racial biases in trained NLP systems, other work (e.g.~included in `Social Science/Social Media' in \Tref{tab:task_overview}) uses NLP tools to analyze race in society. Examples include examining how commentators describe football players of different races \citep{merullo-etal-2019-investigating} or how words like `prejudice' have changed meaning over time \cite{vylomova-etal-2019-evaluation}.

While differing in goals, this work is often susceptible to the same pitfalls as other NLP tasks.
One area requiring particular caution is in the interpretation of results produced by analysis models. For example, while word embeddings have become a common way to measure semantic change or estimate word meanings \cite{garg2018word}, \citet{joseph-morgan-2020-word} show that embedding associations do not always correlate with human opinions; in particular, correlations are stronger for beliefs about gender than race. Relatedly, in HCI, the recognition that authors' own biases can affect their interpretations of results has caused some authors to provide self-disclosures \cite{Schlesinger2017}, but this practice is uncommon in NLP.

We conclude this section by observing that when researchers have looked for racial biases in NLP systems, they have usually found them. This literature calls for proactive approaches in considering how data is collected, annotated, used, and interpreted to prevent NLP systems from exacerbating historical racial hierarchies.

\section{Limitations in where and how NLP operationalizes race}
\label{sec:limitations}

While \Sref{sec:survey} demonstrates ways that NLP systems encode racial biases, we next identify gaps and limitations in how these works have examined racism, focusing on  \textit{how} and \textit{in what tasks} researchers have considered race.
We ultimately conclude that prior NLP literature has marginalized research on race and encourage deeper engagement with other fields, critical views of simplified classification schema, and broader application scope in future work \citep{blodgett-etal-2020-language,hanna2020towards}.

\subsection{Common data sets are narrow in scope}

The papers we surveyed suggest that research on race in NLP has used a very limited range of data sets, which fails to account for the multi-dimensionality of race and simplifications inherent in classification.
We identified 3 common data sources:\footnote{We provide further counts of what racial categories papers use and how they operationalize them in \Aref{sec:appendix_extra_stats}.}

\begin{compactitem}
    \item 9 papers use a set of tweets with inferred probabilistic topic labels based on alignment with U.S. census race/ethnicity groups (or the provided inference model) \cite{blodgett-etal-2016-demographic}.
    \item 11 papers use lists of names drawn from \citet{sweeney2013discrimination}, \citet{Caliskan183}, or \citet{garg2018word}. Most commonly, 6 papers use African/European American names from the Word Embedding Association Test (WEAT) \citep{Caliskan183}, which in turn draws data from \citet{greenwald1998measuring} and \citet{Bertrand2004}.
    \item 10 papers use explicit keywords like `Black woman', often placed in templates like ``I am a \rule{1cm}{0.15mm}'' to test if model performance remains the same for different identity terms.
\end{compactitem}

While these commonly-used data sets can identify performance disparities, they only capture a narrow subset of the multiple dimensions of race (\Sref{sec:race_definition}). For example, none of them capture self-identified race. While observed race is often appropriate for examining discrimination and some types of disparities, it is impossible to assess potential harms and benefits of NLP systems without assessing their performance over text generated by and directed to people of different races. The corpus from \citet{blodgett-etal-2016-demographic} does serve as a starting point and forms the basis of most current work assessing performance gaps in NLP models \cite{sap-etal-2019-risk,blodgett-etal-2018-twitter,xia-etal-2020-demoting,xu-etal-2019-privacy,groenwold-etal-2020-investigating}, but even this corpus is explicitly not intended to infer race.

Furthermore, names and hand-selected identity terms are not sufficient for uncovering model bias. \citet{de2019bias} show this in examining gender bias in occupation classification: when overt indicators like names and pronouns are scrubbed from the data, performance gaps and potential allocational harms still remain. Names also generalize poorly. While identity terms can be examined across languages \cite{van-miltenburg-etal-2017-cross}, differences in naming conventions often do not translate, leading some studies to omit examining racial bias in non-English languages \cite{lauscher-glavas-2019-consistently}. Even within English, names often fail to generalize across domains, geographies, and time. For example,  names drawn from the U.S. census generalize poorly to Twitter \citep{wood-doughty-etal-2018-predicting}, and names common among Black and white children were not distinctly different prior to the 1970s \citep{fryer2004causes,sweeney2013discrimination}.

We focus on these 3 data sets as they were most common in the papers we surveyed, but we note that others exist. \citet{preotiuc-pietro-ungar-2018-user} provide a data set of tweets with self-identified race of their authors, though it is little used in subsequent work and focused on demographic prediction, rather than evaluating model performance gaps. Two recently-released data sets \citep{nadeem2020stereoset,nangia-etal-2020-crows} provide crowd-sourced pairs of more- and less-stereotypical text. More work is needed to understand any privacy concerns and the strengths and limitations of these data   \cite{blodgett2021stereotyping}.
Additionally, some papers collect domain-specific data, such as self-reported race in an online community \citep{loveys-etal-2018-cross}, or crowd-sourced annotations of perceived race of football players \citep{merullo-etal-2019-investigating}. While these works offer clear contextualization, it is difficult to use these data sets to address other research questions.\looseness=-1

\subsection{Classification schemes operationalize race as a fixed, single-dimensional U.S.-census label}

Work that uses the same few data sets inevitably also uses the same few classification schemes, often without justification. The most common explicitly stated source of racial categories is the U.S. census,
which reflects the general trend of U.S.-centrism in NLP research (the vast majority of work we surveyed also focused on English). While census categories are sometimes appropriate, repeated use of classification schemes and accompanying data sets without considering who defined these schemes and whether or not they are appropriate for the current context risks perpetuating the misconception that race is `natural' across geo-cultural contexts. We refer to \citet{hanna2020towards} for a more thorough overview of the harms of ``widespread uncritical adoption of racial categories,'' which ``can in turn re-entrench systems of racial stratification which give rise to real health and social inequalities.'' At best, the way race has been operationalized in NLP research is only capable of examining a narrow subset of potential harms. At worst, it risks reinforcing racism by presenting racial divisions as natural, rather than the product of social and historical context \citep{bowker2000sorting}.

As an example of questioning who devised racial categories and for what purpose, we consider the pattern of re-using names from \citet{greenwald1998measuring}, who describe their data as sets of names ``judged by introductory psychology students to be more likely to belong to White Americans than to Black Americans'' or vice versa.
When incorporating this data into WEAT, \citet{Caliskan183} discard some judged African American names as too infrequent in their embedding data. Work subsequently drawing from WEAT makes no mention of the discarded names  nor contains much discussion of how the data was generated and whether or not names judged to be white or Black by introductory psychology students in 1998 are an appropriate benchmark for the studied task.
While gathering data to examine race in NLP is challenging, and in this work we ourselves draw from examples that use \citet{greenwald1998measuring}, it is difficult to interpret what implications arise when models exhibit disparities over this data and to what extent models without disparities can be considered `debiased'.

Finally, almost all of the work we examined conducts single-dimensional analyses, e.g.~focus on race or gender but not both simultaneously. This focus contrasts with the concept of \textit{intersectionality}, which has shown that examining discrimination along a single axis fails to capture the experiences of people who face marginalization along multiple axes. For example, consideration of race often emphasizes the experience of gender-privileged people (e.g. Black men), while consideration of gender emphasizes the experience of race-privileged people (e.g. white women). Neither reflect the experience of people who face discrimination along both axes (e.g. Black women) \cite{crenshaw1989demarginalizing}. A small selection of papers have examined intersectional biases in embeddings or word co-occurrences \cite{herbelot-etal-2012-distributional,may-etal-2019-measuring,tan2019assessing,lepori-2020-unequal},
but we did not identify mentions of intersectionality in any other NLP research areas. Further, several of these papers use NLP technology to examine or validate theories on intersectionality; they do not draw from theory on intersectionality to critically examine NLP models. These omissions can mask harms: \citet{jiang-fellbaum-2020-interdependencies} provide an example using word embeddings of how failing to consider intersectionality can render invisible people marginalized in multiple ways. Numerous directions remain for exploration, such as how `debiasing' models along one social dimension affects other dimensions. Surveys in HCI offer further frameworks on how to incorporate identity and intersectionality into computational research \citep{Schlesinger2017,Rankin2019}.

\subsection{NLP research on race is restricted to specific tasks and applications}
Finally, \Tref{tab:task_overview} reveals many common NLP applications where race has not been examined, such as machine translation, summarization, or question answering.\footnote{We identified only $8$ relevant papers on Text Generation, which focus on other areas including chat bots, GPT-$2$/$3$, humor generation, and story generation. }
While some tasks seem inherently more relevant to social context than others (a claim we dispute in this work, particularly in \Sref{sec:people}), \textit{research on race is compartmentalized to limited areas of NLP even in comparison with work on `bias'}. For example, \citet{blodgett-etal-2020-language} identify $20$ papers that examine bias in co-reference resolution systems and $8$ in machine translation, whereas we identify $0$ papers in either that consider race.
Instead, race is most often mentioned in NLP papers in the context of abusive language, and work on detecting or removing bias in NLP models has focused on word embeddings.

Overall, our survey identifies a need for the examination of race in a broader range of NLP tasks, the development of multi-dimensional data sets, and careful consideration of context and appropriateness of racial categories. 
In general, race is difficult to operationalize, but NLP researchers do not need to start from scratch, and can instead draw from relevant work in other fields.

\section{NLP propagates marginalization of racialized people}
\label{sec:people}
While in \Sref{sec:limitations} we primarily discuss race as a topic or a construct, in this section, we consider the role, or more pointedly, the absence, of traditionally underrepresented people in NLP research.

\subsection{People create data} As discussed in \Sref{sec:nlp_pipeline}, data and annotations are generated by people, and failure to consider who created data can lead to harms. In \Sref{sec:nlp_pipeline} we identify a need for diverse training data in order to ensure models work for a diverse set of people, and in \Sref{sec:limitations} we describe a similar need for diversity in data that is used to assess algorithmic fairness. However, gathering this type of data without consideration of the people who generated it can introduce privacy violations and risks of demographic profiling.

As an example, in 2019, partially in response to research showing that facial recognition algorithms perform worse on darker-skinned than lighter-skinned people \cite{buolamwini2018gender,raji2019actionable}, researchers at IBM created the ``Diversity in Faces'' data set, which consists of 1 million photos sampled from the the publicly available
YFCC-100M data set and annotated with ``craniofacial distances, areas and ratios, facial
symmetry and contrast, skin color, age and gender predictions'' \citep{merler2019diversity}. While this data set aimed to improve the fairness of facial recognition technology, it included photos collected from a Flickr, a photo-sharing website whose users did not explicitly consent for this use of their photos. Some of these users filed a lawsuit against IBM, in part for ``subjecting them to increased surveillance, stalking, identity theft, and other invasions of privacy and fraud.''\footnote{\url{https://www.classaction.org/news/class-action-accuses-ibm-of-flagrant-violations-of-illinois-biometric-privacy-law-to-develop-facial-recognition-tech\#embedded-document} \\
\url{https://www.nbcnews.com/tech/internet/facial-recognition-s-dirty-little-secret-millions-online-photos-scraped-n981921}
IBM has since removed the ``Diversity in Faces'' data set as well as their ``Detect Faces'' public API and stopped their use of and research on facial recognition. \url{https://qz.com/1866848/why-ibm-abandoned-its-facial-recognition-program/}}
NLP researchers could easily repeat this incident, for example, by using demographic profiling of social media users to create more diverse data sets.
While obtaining diverse, representative, real-world data sets is important for building models, data must be collected with consideration for the people who generated it, such as obtaining informed consent, setting limits of uses, and preserving privacy, as well as recognizing that some communities may not want their data used for NLP at all \citep{paullada2020how}.

\subsection{People build models}
Research is additionally carried out by people who determine what projects to pursue and how to approach them.
While statistics on ACL conferences and publications have focused on geographic representation rather than race, they do highlight under-representation.
Out of $2,695$ author affiliations associated with papers in the ACL Anthology for $5$ major conferences held in $2018$, only $5$ ($0.2$\%) were from Africa, compared with $1,114$ from North America ($41.3$\%).\footnote{\url{http://www.marekrei.com/blog/geographic-diversity-of-nlp-conferences/}} Statistics published for $2017$ conference attendees and ACL fellows similarly reveal a much higher percentage of people from ``North, Central and South America'' ($55$\% attendees / $74$\% fellows) than from ``Europe, Middle East and Africa'' ($19$\%/$13$\%)  or ``Asia-Pacific'' ($23$\%/$13$\%).\footnote{\url{https://www.aclweb.org/portal/content/acl-diversity-statistics}} These broad regional categories likely mask further under-representation, e.g. percentage of attendees and fellows from Africa as compared to Europe. 
According to an NSF report that includes racial statistics rather than nationality, $~14\%$ of doctorate degrees in Computer Science awarded by U.S. institutions to U.S. citizens and permanent residents were awarded to Asian students, $<4$\% to Black or African American students, and 0\% to American Indian or Alaska Native students  \cite{NSF2019}.\footnote{Results exclude respondents who did not report race or ethnicity or were Native Hawaiian or Other Pacific Islander.}

It is difficult to envision reducing or eliminating racial differences in NLP systems without changes in the researchers building these systems. One theory that exemplifies this challenge is \textit{interest convergence}, which suggests that people in positions of power only take action against systematic problems like racism when it also advances their own interests \citep{bell1980brown}.
\citet{Ogbonnaya-Ogburu2020} identify instances of interest convergence in the HCI community, primarily in diversity initiatives that benefit institutions' images rather than underrepresented people.
In a research setting, interest convergence can encourage studies of incremental and surface-level biases while discouraging research that might be perceived as controversial and force fundamental changes in the field. 

Demographic statistics are not sufficient for avoiding pitfalls like interest convergence, as they fail to capture the lived experiences of researchers. \citet{Ogbonnaya-Ogburu2020} provide several examples of challenges that non-white HCI researchers have faced, including the invisible labor of representing `diversity', everyday microaggressions, and altering their research directions in accordance with their advisors' interests. \citet{Rankin2019} further discuss how research conducted by people of different races is perceived differently: ``Black women in academia who conduct research about the intersections of race, gender, class, and so on are perceived as `doing service,' whereas white colleagues who conduct the same research are perceived as doing cutting-edge research that demands attention and recognition.'' While we draw examples about race from HCI in the absence of published work on these topics in NLP, the lack of linguistic diversity in NLP research similarly demonstrates how representation does not necessarily imply inclusion. Although researchers from various parts of the world (Asia, in particular) do have some numerical representation among ACL authors, attendees, and fellows, NLP research overwhelmingly favors a small set of languages, with a heavy skew towards European languages \citep{joshi-etal-2020-state} and `standard' language varieties \citep{kumar-etal-2021-langvar}.


\subsection{People use models}
Finally, NLP research produces technology that is used by people, and even work without direct applications is typically intended for incorporation into application-based systems. 
With the recognition that technology ultimately affects people, researchers on ethics in NLP have increasingly called for considerations of whom technology might harm and suggested that there are some NLP technologies that should not be built at all.
In the context of perpetuating racism, examples include criticism of tools for predicting demographic information \cite{Tatman2020} and automatic prison term prediction \cite{leins-etal-2020-give}, motivated by the history of using technology to police racial minorities and related criticism in other fields \cite{browne2015dark,buolamwini2018gender,mcilwain2019black}. In cases where potential harms are less direct, they are often unaddressed entirely. For example, while low-resource NLP is a large area of research, a paper on machine translation of white American and European languages is unlikely to discuss how continual model improvements in these settings increase technological inequality. Little work on low-resource NLP has focused on the realities of structural racism or differences in lived experience and how they might affect the way technology should be designed.

Detection of abusive language offers an informative case study on the danger of failing to consider people affected by technology. Work on abusive language often aims to detect racism for content moderation \cite{waseem-hovy-2016-hateful}. However, more recent work has show that existing hate speech classifiers are likely to falsely label text containing identity terms like `black' or text containing linguistic markers of \AAE as toxic \cite{dixon2018measuring,sap-etal-2019-risk,davidson-etal-2019-racial,xia-etal-2020-demoting}. Deploying these models could censor the posts of the very people they purport to help.

In other areas of statistics and machine learning, focus on \textit{participatory design} has sought to amplify the voices of people affected by technology and its development. An ICML 2020 workshop titled ``Participatory Approaches to Machine Learning''  highlights a number of papers in this area 
\cite{paml2020,Brown2019}. A few related examples exist in NLP, e.g.~\citet{gupta-etal-2020-heart} gather data for an interactive dialogue agent intended to provide more accessible information about heart failure to Hispanic/Latinx and African American patients.
The authors engage with healthcare providers and doctors, though they leave focal groups with patients for future work. While NLP researchers may not be best situated to examine how people interact with deployed technology, they could instead draw motivation from fields that have stronger histories of participatory design, such as HCI. However, we did not identify citing participatory design studies conducted by others as common practice in the work we surveyed. As in the case of researcher demographics,  participatory design is not an end-all solution. \citet{sloane2020participation} provide a discussion of how participatory design can collapse to `participation-washing' and how such work must be context-specific, long-term, and genuine.

\section{Discussion}
We conclude by synthesizing some of the observations made in the preceding sections into more actionable items.
First, NLP research needs to explicitly incorporate race. We quote \citet{benjamin2019race}: 
\textit{“[technical systems and social codes] operate within powerful systems of meaning that render some things visible, others invisible, and create a vast array of distortions and dangers.”}

In the context of NLP research, this philosophy implies that all technology we build works in service of some ideas or relations, either by upholding them or dismantling them. Any research that is not actively combating prevalent social systems like racism risks perpetuating or exacerbating them. Our work identifies several ways in which NLP research upholds racism:

\begin{compactitem}
    \item Systems contain representational harms and performance gaps throughout NLP pipelines
    \item Research on race is restricted to a narrow subset of tasks and definitions of race, which can mask harms and falsely reify race as `natural'
    \item Traditionally underrepresented people are excluded from the research process, both as consumers and producers of technology
\end{compactitem}

Furthermore, while we focus on race, which we note has received substantially less attention than gender, many of the observations in this work hold for social characteristics that have received even less attention in NLP research, such as socioeconomic class, disability, or sexual orientation \cite{Mendelsohn2020,hutchinson-etal-2020-social}.

Nevertheless, none of these challenges can be addressed without direct engagement with marginalized communities of color. NLP researchers can draw on precedents for this type of engagement from other fields, such as participatory design and value sensitive design models \citep{Friedman2006}. Additionally, numerous organizations already exist that serve as starting points for partnerships, such as \href{https://blackinai.github.io/}{Black in AI}, \href{https://www.masakhane.io/}{Masakhane}, \href{https://d4bl.org/}{Data for Black Lives}, and the \href{https://www.ajl.org/}{Algorithmic Justice League}.

Finally, race and language are complicated, and while readers may look for clearer recommendations, no one data set, model, or set of guidelines can `solve' racism in NLP. For instance, while we draw from linguistics, \citet{hudley2020toward} in turn call on linguists to draw models of racial justice from anthropology, sociology, and psychology. Relatedly, there are numerous racialized effects that NLP research can have that we do not address in this work; for example, \citet{bender-2021} and \citet{strubell-etal-2019-energy} discuss the environmental costs of training large language models, and how global warming disproportionately affects marginalized communities. We suggest that readers use our work as one starting point for bringing inclusion and racial justice into NLP.

\section*{Acknowledgements}
We gratefully thank Hanna Kim, Kartik Goyal, Artidoro Pagnoni, Qinlan Shen, and Michael Miller Yoder for their feedback on this work.
Z.W. has been supported in part by the Canada 150 Research Chair program and the UK-Canada Artificial Intelligence Initiative.
A.F. has been supported in part by a Google PhD Fellowship and a GRFP under Grant No.~DGE1745016.
This material is based upon work supported in part by the National Science Foundation under Grants No.~IIS2040926 and IIS2007960. Any opinions, findings, and conclusions or recommendations expressed in this material are those of the authors and do not necessarily reflect the views of the NSF.

\section{Ethical Considerations}
We, the authors of this work, are situated in the cultural contexts of the United States of America and the United Kingdom/Europe, and some of us identify as people of color. We all identify as NLP researchers, and we acknowledge that we are situated within the traditionally exclusionary practices of academic research. These perspectives have impacted our work, and there are viewpoints outside of our institutions and experiences that our work may not fully represent.

\bibliography{anthology,acl2020}
\bibliographystyle{acl_natbib}

\clearpage
\setcounter{page}{1}
\appendix

\section{ACL Anthology Venues}
\label{sec:appendix_venues}
ACL events: AACL, ACL, ANLP, CL, CoNLL, EACL, EMNLP, Findings, NAACL, SemEval, *SEM, TACL, WMT, Workshops, Special Interest Groups

Non-ACL events: ALTA, AMTA, CCL, COLING, EAMT, HLT, IJCNLP, JEP/TALN/RECITAL, LILT, LREC, MUC, PACLIC, RANLP, ROCLING/IJCLCLP, TINLAP, TIPSTER

\section{Additional Survey Metrics}
\label{sec:appendix_extra_stats}

\begin{figure}
    \centering
    \includegraphics[width=\columnwidth]{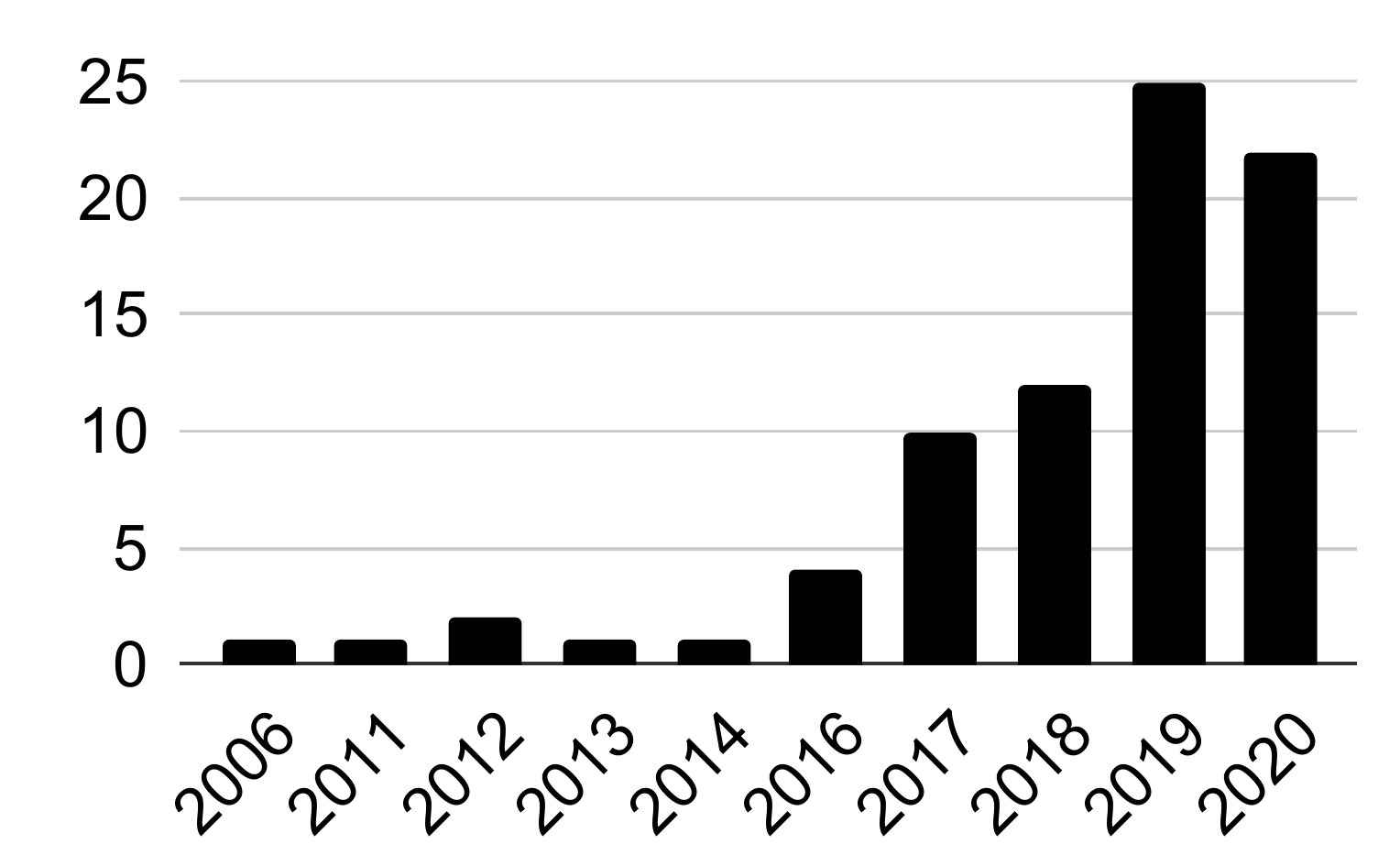}
    \caption{Year of publication of $79$ papers that mention ``racial'' or ``racism''. More papers have been published in recent years (2019-2020).}
    \label{fig:yearHist}
\end{figure}

\begin{figure}
    \centering
    \includegraphics[width=\columnwidth]{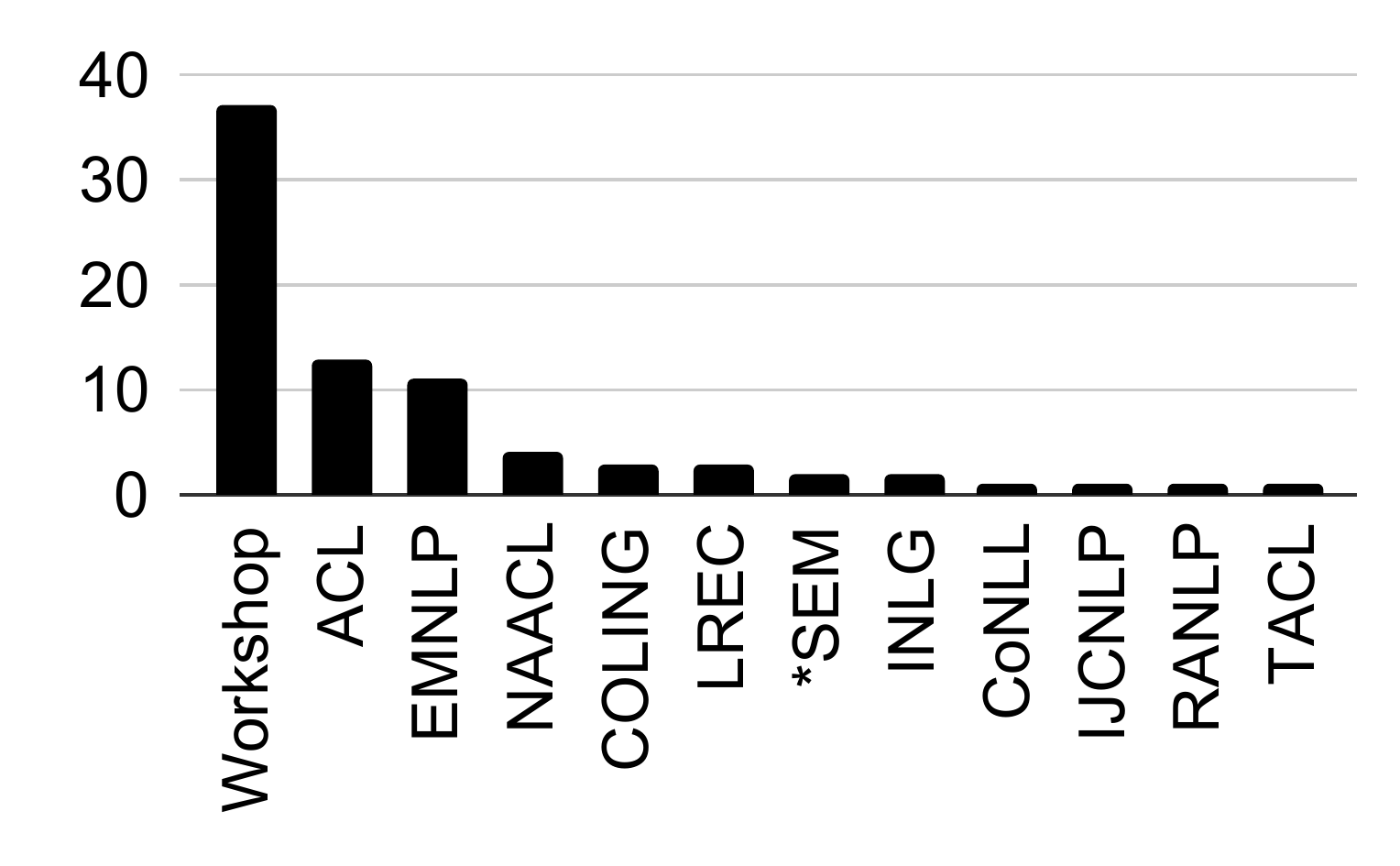}
    \caption{Venue of publication of $79$ papers that mention ``racial'' or ``racism''. About half (46.8\%) were published in workshops.}
    \label{fig:venueHist}
\end{figure}

We show three additional breakdowns of the data set: \Fref{fig:yearHist} shows the number of papers published each year, \Fref{fig:venueHist} shows the number of papers published in each venue, and \Tref{tab:race_categories} shows how papers have operationalized race. As expected, given the growth of NLP research in general and the increasing focus on social issues (e.g. ``Ethics and NLP'' track was added to ACL in 2020) more work has been published on race in more recent years (2019, 2020). In \Fref{fig:venueHist}, we consider if work on race has been siloed into or out of specific venues. The majority of papers were published in workshops, which is consist with the large number of workshop papers. In 2019, approximately 2,038 papers were published in workshops\footnote{\url{https://www.aclweb.org/anthology/venues/ws/}} and 1,680 papers were published in conferences (ACL, EMNLP, NAACL, CONLL, CICLing), meaning 54.8\% were published in workshops. In our data set, 46.8\% of papers surveyed were published in workshops. The most number of papers were published in the largest conferences: ACL and EMNLP. Thus, while \Tref{tab:task_overview} suggests that discussions of race have been siloed to particular NLP applications, \Fref{fig:venueHist} does not show evidence that they have been siloed to particular venues.

\begin{table}
    \centering
    \resizebox{7.5cm}{!}{
    \begin{tabular}{ccccccccc}
& \rotatebox[origin=l]{90}{Census-aligned} & \rotatebox[origin=l]{90}{Crowd-sourced} & \rotatebox[origin=l]{90}{Explicit keywords} & \rotatebox[origin=l]{90}{External/Public} & \rotatebox[origin=l]{90}{Names} & \rotatebox[origin=l]{90}{Predicted} & \rotatebox[origin=l]{90}{Self-reported} & Total \\
\hline
\hline
4+ &  &  & 5 & 2 &  & 1 & 5 & 13 \\
BW & 7 &  & 2 & 1 & 8 & 1 & 1 & 20 \\
BWAH & 1 &  &  &  &  & 3 &  & 4 \\
\{BWAH\} & 1 & 1 & 3 & 1 & 2 &  &  & 8 \\
W/non-W &  & 1 &  &  & 1 &  &  & 2 \\
\hline
Total & 9 & 2 & 10 & 4 & 11 & 5 & 6 & 47 \\
    \end{tabular}
    }
    \caption{Racial categories used by ACL Anthology papers. BWAH stand for Black, White, Asian, and Hispanic. \{BWAH\} denotes any incomplete subset of BWAH other than BW (e.g. Black and Hispanic). 4+ denotes that the paper used $\ge4$ racial categories, often including ``other'', ``mixed'', or an open-ended text box. Papers with multiple schema are counted as separate data points.}
    \label{tab:race_categories}
\end{table}

In \Tref{tab:race_categories}, for all papers that use categorization schema to classify race, we show what racial categories they use. If a paper uses multiple schemes (e.g.~collects crowd-sourced annotations of stereotypes associated with different races and also asks annotators to self-report their race), we report each scheme as a separate data point. This table does not include papers that do not specify racial categories (e.g.~examine ``racist language'' without specifying targeted people or analyze semantic change of topics like ``racism'' and ``prejudice''). Finally, we map terms used by papers to the ones in \Tref{tab:race_categories}, e.g.~papers examining African American vs. European American names are included in BW.

The majority of papers focus on binary Black/white racial categories. While many papers draw definitions from the U.S.~census, very few papers consider less-commonly-selected census categories like Native American or Pacific Islander. The most common method for identifying people's race uses first or last names (10 papers) or explicit keywords like ``black'' and ``white'' (10 papers).

\clearpage
\onecolumn
\section{Full List of Surveyed Papers}
\label{sec:appendix_papers}

\begin{longtable}{lccccccc}
\hline
 & Year & Venue & NLP Task & Task Type \\
    \hline
    \hline
\footnotesize{\citet{assimakopoulos-etal-2020-annotating}} & 2020 & LREC & Abusive Language & Collect Corpus \\
\footnotesize{\citet{bommasani-etal-2020-interpreting}} & 2020 & ACL & Text Representations & Detect Bias \\
\footnotesize{\citet{chakravarthi-2020-hopeedi}} & 2020 & Workshop & Abusive Language & Collect Corpus \\
\footnotesize{\citet{groenwold-etal-2020-investigating}} & 2020 & EMNLP & Text Generation & Detect Bias \\
\footnotesize{\citet{gupta-etal-2020-heart}} & 2020 & Workshop & Sector-spec. NLP apps. & Collect Corpus \\
\footnotesize{\citet{huang-etal-2020-multilingual}} & 2020 & LREC & Abusive Language & Detect Bias \\
\footnotesize{\citet{jiang-fellbaum-2020-interdependencies}} & 2020 & Workshop & Text Representations & Detect Bias \\
\footnotesize{\citet{joseph-morgan-2020-word}} & 2020 & ACL & Text Representations & Detect Bias \\
\footnotesize{\citet{kennedy-etal-2020-contextualizing}} & 2020 & ACL & Abusive Language & Debias \\
\footnotesize{\citet{kurrek-etal-2020-towards}} & 2020 & Workshop & Abusive Language & Collect Corpus \\
\footnotesize{\citet{lepori-2020-unequal}} & 2020 & COLING & Text Representations & Detect Bias \\
\footnotesize{\citet{liu-etal-2020-gender}} & 2020 & COLING & Text Generation & Debias \\
\footnotesize{\citet{meaney-2020-crossing}} & 2020 & Workshop & Social Science/Media & Survey/Position \\
\footnotesize{\citet{nangia-etal-2020-crows}} & 2020 & EMNLP & Text Representations & Detect Bias \\
\footnotesize{\citet{roy-goldwasser-2020-weakly}} & 2020 & EMNLP & Social Science/Media & Analyze Corpus \\
\footnotesize{\citet{sap-etal-2020-social}} & 2020 & ACL & Abusive Language & Collect Corpus \\
\footnotesize{\citet{shah-etal-2020-predictive}} & 2020 & ACL & Ethics/Task-indep. Bias & Survey/Position \\
\footnotesize{\citet{shahid-etal-2020-detecting}} & 2020 & Workshop & Social Science/Media & Analyze Corpus \\
\footnotesize{\citet{tan-etal-2020-morphin}} & 2020 & ACL & Ethics/Task-indep. Bias & Develop Model \\
\footnotesize{\citet{xia-etal-2020-demoting}} & 2020 & Workshop & Abusive Language & Debias \\
\footnotesize{\citet{zhang-etal-2020-demographics}} & 2020 & ACL & Abusive Language & Detect Bias \\
\footnotesize{\citet{zhao-chang-2020-logan}} & 2020 & EMNLP & Ethics/Task-indep. Bias & Detect Bias \\
\footnotesize{\citet{amir-etal-2019-mental}} & 2019 & Workshop & Sector-spec. NLP apps. & Analyze Corpus \\
\footnotesize{\citet{davidson-etal-2019-racial}} & 2019 & Workshop & Abusive Language & Detect Bias \\
\footnotesize{\citet{demszky-etal-2019-analyzing}} & 2019 & NAACL & Social Science/Media & Analyze Corpus \\
\footnotesize{\citet{gillani-levy-2019-simple}} & 2019 & Workshop & Text Representations & Analyze Corpus \\
\footnotesize{\citet{jurgens-etal-2019-just}} & 2019 & ACL & Abusive Language & Survey/Position \\
\footnotesize{\citet{karve-etal-2019-conceptor}} & 2019 & Workshop & Text Representations & Debias \\
\footnotesize{\citet{kurita-etal-2019-measuring}} & 2019 & Workshop & Text Representations & Detect Bias \\
\footnotesize{\citet{lauscher-glavas-2019-consistently}} & 2019 & Workshop & Text Representations & Detect Bias \\
\footnotesize{\citet{lee-etal-2019-exploring}} & 2019 & Workshop & Text Generation & Detect Bias \\
\footnotesize{\citet{liu-etal-2019-detecting}} & 2019 & CoNLL & Social Science/Media & Develop Model \\
\footnotesize{\citet{manzini-etal-2019-black}} & 2019 & NAACL & Text Representations & Debias \\
\footnotesize{\citet{may-etal-2019-measuring}} & 2019 & ACL & Text Representations & Detect Bias \\
\footnotesize{\citet{mayfield-etal-2019-equity}} & 2019 & Workshop & Sector-spec. NLP apps. & Survey/Position \\
\footnotesize{\citet{merullo-etal-2019-investigating}} & 2019 & EMNLP & Social Science/Media & Analyze Corpus \\
\footnotesize{\citet{mostafazadeh-davani-etal-2019-reporting}} & 2019 & EMNLP & Core NLP Applications & Develop Model \\
\footnotesize{\citet{parish-morris-2019-computational}} & 2019 & Workshop & Sector-spec. NLP apps. & Survey/Position \\
\footnotesize{\citet{romanov-etal-2019-whats}} & 2019 & NAACL & Sector-spec. NLP apps. & Debias \\
\footnotesize{\citet{santos-paraboni-2019-moral}} & 2019 & RANLP & Social Science/Media & Collect Corpus \\
\footnotesize{\citet{sap-etal-2019-risk}} & 2019 & ACL & Abusive Language & Detect Bias \\
\footnotesize{\citet{sharifirad-matwin-2019-using}} & 2019 & Workshop & Abusive Language & Analyze Corpus \\
\footnotesize{\citet{sommerauer-fokkens-2019-conceptual}} & 2019 & Workshop & Text Representations & Detect Bias \\
\footnotesize{\citet{tripodi-etal-2019-tracing}} & 2019 & Workshop & Text Representations & Analyze Corpus \\
\footnotesize{\citet{vylomova-etal-2019-evaluation}} & 2019 & Workshop & Social Science/Media & Analyze Corpus \\
\footnotesize{\citet{wallace-etal-2019-universal}} & 2019 & EMNLP & Text Generation & Detect Bias \\
\footnotesize{\citet{xu-etal-2019-privacy}} & 2019 & INLG & Text Generation & Develop Model \\
\footnotesize{\citet{barbieri-camacho-collados-2018-gender}} & 2018 & *SEM & Social Science/Media & Analyze Corpus \\
\footnotesize{\citet{blodgett-etal-2018-twitter}} & 2018 & ACL & Core NLP Applications & Debias \\
\footnotesize{\citet{castelle-2018-linguistic}} & 2018 & Workshop & Abusive Language & Analyze Corpus \\
\footnotesize{\citet{de-gibert-etal-2018-hate}} & 2018 & Workshop & Abusive Language & Collect Corpus \\
\footnotesize{\citet{elazar-goldberg-2018-adversarial}} & 2018 & EMNLP & Ethics/Task-indep. Bias & Debias \\
\footnotesize{\citet{kasunic-kaufman-2018-learning}} & 2018 & Workshop & Text Generation & Survey/Position \\
\footnotesize{\citet{kiritchenko-mohammad-2018-examining}} & 2018 & *SEM & Social Science/Media & Detect Bias \\
\footnotesize{\citet{loveys-etal-2018-cross}} & 2018 & Workshop & Sector-spec. NLP apps. & Analyze Corpus \\
\footnotesize{\citet{preotiuc-pietro-ungar-2018-user}} & 2018 & COLING & Social Science/Media & Develop Model \\
\footnotesize{\citet{sheng-etal-2019-woman}} & 2018 & EMNLP & Text Generation & Detect Bias \\
\footnotesize{\citet{wojatzki-etal-2018-quantifying}} & 2018 & LREC & Social Science/Media & Collect Corpus \\
\footnotesize{\citet{wood-doughty-etal-2018-predicting}} & 2018 & Workshop & Social Science/Media & Develop Model \\
\footnotesize{\citet{clarke-grieve-2017-dimensions}} & 2017 & Workshop & Abusive Language & Analyze Corpus \\
\footnotesize{\citet{gallagher-etal-2017-anchored}} & 2017 & TACL & Social Science/Media & Develop Model \\
\footnotesize{\citet{hasanuzzaman-etal-2017-demographic}} & 2017 & IJCNLP & Abusive Language & Develop Model \\
\footnotesize{\citet{ramakrishna-etal-2017-linguistic}} & 2017 & ACL & Social Science/Media & Analyze Corpus \\
\footnotesize{\citet{rudinger-etal-2017-social}} & 2017 & Workshop & Core NLP Applications & Detect Bias \\
\footnotesize{\citet{schnoebelen-2017-goal}} & 2017 & Workshop & Ethics/Task-indep. Bias & Survey/Position \\
\footnotesize{\citet{van-miltenburg-etal-2017-cross}} & 2017 & INLG & Image Processing & Detect Bias \\
\footnotesize{\citet{waseem-etal-2017-understanding}} & 2017 & Workshop & Abusive Language & Survey/Position \\
\footnotesize{\citet{wood-doughty-etal-2017-twitter}} & 2017 & Workshop & Social Science/Media & Analyze Corpus \\
\footnotesize{\citet{wright-etal-2017-vectors}} & 2017 & Workshop & Abusive Language & Analyze Corpus \\
\footnotesize{\citet{blodgett-etal-2016-demographic}} & 2016 & EMNLP & Ethics/Task-indep. Bias & Collect Corpus \\
\footnotesize{\citet{pavlick-etal-2016-gun}} & 2016 & EMNLP & Core NLP Applications & Collect Corpus \\
\footnotesize{\citet{waseem-2016-racist}} & 2016 & Workshop & Abusive Language & Detect Bias \\
\footnotesize{\citet{waseem-hovy-2016-hateful}} & 2016 & Workshop & Abusive Language & Collect Corpus \\
\footnotesize{\citet{mohammady-culotta-2014-using}} & 2014 & Workshop & Social Science/Media & Develop Model \\
\footnotesize{\citet{bergsma-etal-2013-broadly}} & 2013 & NAACL & Social Science/Media & Develop Model \\
\footnotesize{\citet{herbelot-etal-2012-distributional}} & 2012 & Workshop & Social Science/Media & Analyze Corpus \\
\footnotesize{\citet{warner-hirschberg-2012-detecting}} & 2012 & Workshop & Abusive Language & Develop Model \\
\footnotesize{\citet{eisenstein-etal-2011-discovering}} & 2011 & ACL & Social Science/Media & Analyze Corpus \\
\footnotesize{\citet{somers-2006-language}} & 2006 & Workshop & Sector-spec. NLP apps. & Survey/Position \\
\end{longtable}
\clearpage
\twocolumn

\end{document}